\documentclass{article}
\usepackage{booktabs}
\usepackage{multirow}
\usepackage{spconf,amsmath,graphicx}
\usepackage{amsmath, amsthm, amssymb}
\usepackage{enumerate}
\usepackage{xfrac}
\usepackage{xcolor}
\usepackage{multicol}
\usepackage{amsmath}

\usepackage {subcaption}
\usepackage{graphicx}
\usepackage{algorithm, algorithmic}
\usepackage{mathrsfs}
\usepackage{bbm}
\usepackage{xspace}
\usepackage{natbib}
\setcitestyle{numbers,square}

\usepackage{hyperref}
\hypersetup{
    colorlinks=true,
    linkcolor=blue,
    filecolor=blue,
    urlcolor=blue,
    citecolor=cyan,
}

\newcommand{\R}{\mathbb{R}}

\newcommand{\eps}{\varepsilon}
\renewcommand{\epsilon}{\eps}

\title{Wavelet-Inspired Multiscale Graph Convolutional Recurrent Network for Traffic Forecasting}
\name{Qipeng Qian $^{1,2}$, Tanwi Mallick $^{2}$
}
\address{$^{1}$ Department of Applied Mathematics, University of Arizona, Tucson, AZ, USA\\
$^{2}$ Mathematics and Computer Science Division, Argonne National Laboratory, IL, USA }

\begin{document}
%
\maketitle
\begin{abstract}
Traffic forecasting is the foundation for intelligent transportation systems. Spatiotemporal graph neural networks have demonstrated state-of-the-art performance in traffic forecasting.
However, these methods do not explicitly model some of the natural characteristics in traffic data, such as the multiscale structure that encompasses spatial and temporal variations at different levels of granularity or scale.
To that end, we propose a Wavelet-Inspired Graph Convolutional Recurrent Network (WavGCRN)  which combines multiscale analysis (MSA)-based method with Deep Learning (DL)-based method.
In WavGCRN, the traffic data is decomposed into time-frequency components with Discrete Wavelet Transformation (DWT), constructing a multi-stream input structure;
then Graph Convolutional Recurrent networks (GCRNs) are employed as encoders for each stream, extracting spatiotemporal features in different scales;
and finally the learnable Inversed DWT and GCRN are combined as the decoder, fusing the information from all streams for traffic metrics reconstruction and prediction.  Furthermore, road-network-informed graphs and data-driven graph learning are combined to accurately capture spatial correlation.
The proposed method can offer well-defined interpretability, powerful learning capability, and competitive forecasting performance on real-world traffic data sets.
\end{abstract}
\begin{keywords}
Spatiotemporal correlation, wavelet-inspired network, graph learning, traffic forecasting
\end{keywords}
\section{Introduction}
\label{sec:intro}

The intelligent transportation system (ITS) has garnered significant attention due to ever-escalating traffic pressures.
In this context, traffic forecasting plays a pivotal role in comprehending and crafting an efficient transportation system while mitigating traffic congestion.

Traditional traffic forecasting methods have their origins in time series analysis, e.g., vector auto-regression, support vector regression, auto-regressive integrated moving average (ARIMA), hidden Markov model. Their model assumptions are frequently at odds with real-world traffic data, and complex domain knowledge is needed, such as queuing theory and behavior simulation in various traffic scenarios \cite{cascetta2013}.
In recent years, Deep Learning (DL)-based methods have emerged as a mainstream trend in traffic forecasting.
They autonomously unearth the underlying information of traffic data, bypassing the need for crafting a traffic metrics model tailor-made for each road segment involved.

The spatiotemporal correlation in traffic data is the basis of modeling and prediction, therefore how to effectively exploit spatiotemporal information in learning procedures is a crucial problem that largely determines the predictive performance.
Graphs are effective tools to represent spatial correlations in distributed observed data, so Graph Neural Networks (GNNs) have been used to
analyze signals or features residing on graph \cite{kipf2016gcn}.
On the other hand, to capture the temporal correlations of traffic data, Recurrent Neural Networks (RNNs) like Long Short-Term Memory (LSTM) \cite{hochreiter1997long}, Gated Recurrent Unit (GRU) \cite{chung2014empirical}, and Transformer \cite{vaswani2017attention} have gained wide recognition.
Integrating GNNs with RNNs leads to a number of state-of-the-art traffic forecasting methods. DCRNN \cite{li2017dcrnn} is the first work to combine GNN and RNN.
STGCN \cite{yu2017spatio} applies CNN instead of RNN to improve the model efficiency at the expense of some performance.
DGCRN \cite{li2023dynamic} considers the dynamic change of the underlying graph.
Hierarchical GCN  \cite{guo2021hierarchical} exploits the hierarchical patterns of traffic data.

However, these DL-based methods can not explicitly investigate some natural characteristics in traffic data, such as multiscale structure, necessitating high-quality data and artful strategies to capture these features. On the other hand, Multiscale Analysis (MSA) models can represent the hidden frequency information through rigors mathematical and physical perspectives \cite{kriechbaumer2014improved}, \cite{wang2018multilevel}, \cite{chen2021multiscale}.
Thus explicitly decomposing traffic data into different frequency domains can decouple the complex relationship and facilitate learning and interpretation.
However, effectively combining MSA with DL into a consistent method is still a challenging problem in traffic forecasting research,
especially coupling multiscale spatiotemporal feature learning and traffic graph learning, designing MSA-based DL architecture, and understanding prediction methods and results \cite{huang2022physics}.

In this paper, we proposed a Wavelet-Inspired Graph Convolutional Recurrent Network (WavGCRN), which is a multi-stream encoder-decoder architecture.
Discrete wavelet transform (DWT) and learnable inverse DWT (LIDWT) are combined with graph convolutional recurrent network (GCRN)
to make up the backbone modules.
This network is MSA-based and DL-based, aligning the wavelet transform process with the deep network architecture.
Meanwhile,  distinct graphs are learned for each stream, making the graphs better reflect the multiscale correlations. The main contributions include:

a) DWT is applied to construct a multi-stream encoder. DWT-based signal decomposition can decouple traffic information, augmenting the power of data-driven DL and helping to uncover the implicit relationship buried in the data.

b) LIDWT is used as a fusion decoder. This architecture not only aggregates learned features from different encoder streams but simulates the transformation and inverse transformation processes of wavelet as well, which will benefit interpretation and implementation.

c) An optimization-based graph learning method is combined with a prior road-network-informed graph to take full advantage of physical-informed and data-driven graph generation methods.

\begin{figure*}[htb]
  \centering
  \includegraphics[width=1.\textwidth]{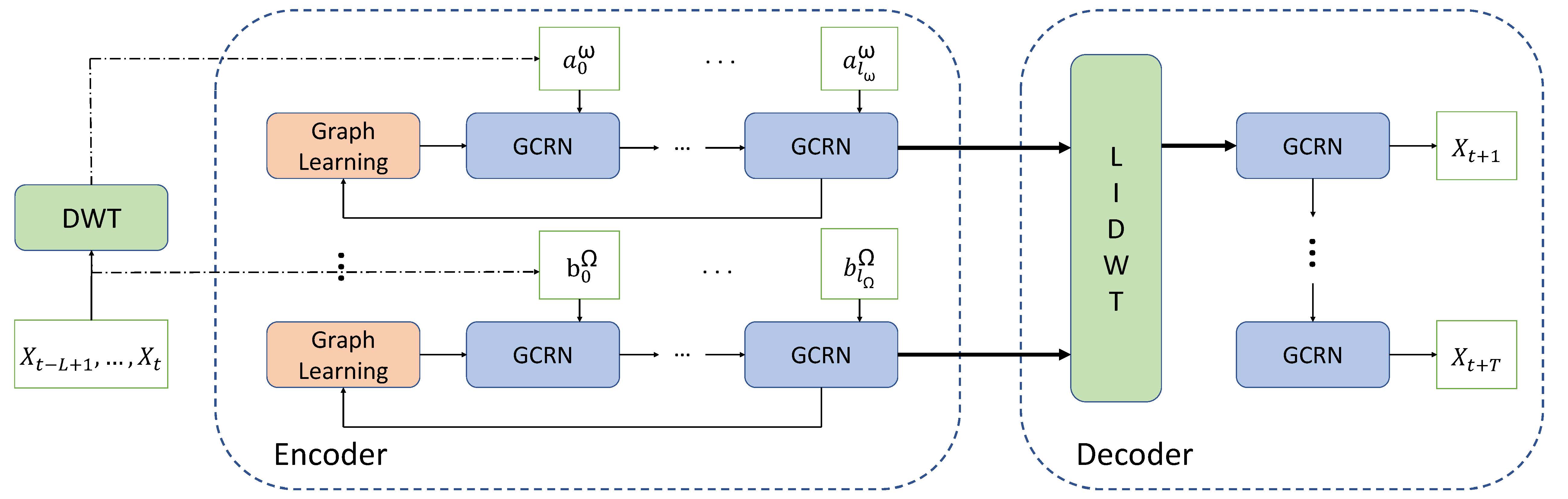}
  \caption{The architecture of WavGCRN with DWT-based multi-stream encoder and LIDWT-based multi-stream fused decoder.
  }
\end{figure*}\label{flowchart}

\section{Methodology}
\label{sec:method}
The traffic forecasting problem discussed in this paper is specifically defined as predicting future traffic metrics given previously observed traffic metrics from $N$ correlated sensors located on the different positions in a road network. We represent the sensor relationship as a weighted directed graph $G=(V,E,A)$ with $|V|=N$. Denote the traffic metrics observed on $G$ as a graph signal $X\in\R^{N\times d}$, where $d$ is the feature dimension for each node. Let $X^t$ represent the graph signal observed at time $t$. The traffic forecasting task is learning a function $h$ that maps the historical data of length $L$ to the future prediction of length $T$:
\begin{equation}
h(X^{t-L+1},...,X^t; G)=(X^{t+1},...,X^{t+T})
\end{equation}

WavGCRN is shown in Figure \ref{flowchart}.
The traffic metrics signal of each sensor is decomposed by DWT and each decomposed component constitutes a stream of encoder.
In each stream, the spatiotemporal feature of each sensor is extracted by a GCRN-based encoder. The encoded features of all streams are fused by LIDWT and then input to a GCRN-based decoder to restore and predict the traffic metrics signals of all sensors. The graph is updated iteratively based on the learned node features in the current training iteration.

\subsection{DWT-based Multi-stream Input}

MSA is a good choice to decouple the spatiotemporal structure by either spatial-multiscale or temporal-multiscale decomposition. In general, the low-frequency signal components are able to capture slow-varying character while the high-frequency ones embody fine-grained changes.
Separating these different scale components, addressing them respectively, and fusing them interactively in a network model, will enhance the model interpretability and reduce learning load.

DWT is a well-known tool for time series analyze \cite{kriechbaumer2014improved}  due to its multiscale time-frequency property.
For raw input time series $X\in\R^{N\times L}$, if Haar wavelet is used, the DWT coefficients at level $\omega$ is calculated as
\begin{equation}
\begin{aligned}
&a^{(\omega)}_{i,:}(k)=\frac{1}{2}[b^{(\omega-1)}_{i,:}(2k)+b^{(\omega-1)}_{i,:}(2k+1)],\\
&b^{(\omega)}_{i,:}(k)=\frac{1}{2}[b^{(\omega-1)}_{i,:}(2k)-b^{(\omega-1)}_{i,:}(2k+1)].
\end{aligned}
\end{equation}
where $a^{(\omega)}$ is approximate coefficients and $b^{(0)}=X$, $b^{(\omega)}$ is detailed coefficients. The output of $\Omega$-level DWT is
$$\{a^{(\omega)}\}_{\omega=1}^{\Omega}\cup\{b^{(\Omega)}\}.$$

\subsection{Graph Convolutional Recurrent Network (GCRN)}

GCRN is built by GCN with GRU  \cite{li2017dcrnn}, \cite{li2023dynamic}, which is used for multi-stream encoder to process each signal component individually, and for fusion decoder to restore and predict traffic metrics.

The message propagation rule of GCN is derived from graph spectral convolution \cite{chung1997spectral}, which works as a parameterized filter $g_{\theta}$ in the frequency domain. Chebyshev spectral CNN \cite{defferrard2016convolutional} uses Chebyshev polynomials to approximate $g_{\theta}$ with predefined order $k$ and with order $1$ we get GCN. We use $K$-hop graph convolution defined as
\begin{equation}
\begin{aligned}
&H^{(k)}=\alpha H_{in} + \beta \Tilde{A}H^{(k-1)},\\
&H_{out}=\sum_{k=0}^KH^{(k)}W^{(k)},\, H^{(0)}=H_{in},\\
&\Tilde{A}=D^{-1} A.
\end{aligned}
\end{equation}
where $A$ is the adjacent matrix of current learned graph, and $D$ is its degree matrix.
$H_{in}$ is the input hidden state, $H_{out}$ is the output of $K$-hop graph convolution.
$W^{(k)}\in\R^{d^{in}\times d^{out}}$ are learnable parameters,  $d^{in}$ and $d^{out}$ are the input and output dimensions.
$\alpha$ and $\beta$ are weight parameters.
Abbreviate the above equations as
$$H_{out}=\Theta*_{G}(H_{in}, A).$$
Replacing matrix multiplication in GRU with $*_{G}$, we get
\begin{equation}
\begin{aligned}
&r_t=\sigma(\Theta_r*_{G}(a^{(\omega)}_{:,t}|H^{(\omega)}_{t-1}, A^{(\omega)})),\\
&u_t=\sigma(\Theta_u*_{G}(a^{(\omega)}_{:,t}|H^{(\omega)}_{t-1}, A^{(\omega)})),\\
&C_t=\tan(\Theta_C*_{G}(a^{(\omega)}_{:,t}|r_t\odot H^{(\omega)}_{t-1}, A^{(\omega)})),\\
&H^{(\omega)}_{t}=u_t\odot C_t+(1-u_t)\odot H^{(\omega)}_{t-1},
\end{aligned}
\end{equation}
where $|$ is the concatenation operation.

For each stream of the encoder, the number of layers formed by GCRN is determined by the input signal length. The number of layers formed by GCRN in decoder is determined by the time horizon of the specific prediction task.

\subsection{Learnable Inverse DWT (LIDWT)}

Under neural network framework, signal transformation can be parameterized/learnable \cite{cirstea2022triformer}, \cite{chen2021multiscale}.
To ensure consistency with the DWT process, we incorporate a LIDWT to fuse the outputs from the multi-stream encoder.
The proposed LIDWT strikes a balance between preserving the physical significance of the wavelet-transform-based representation while benefiting from the expressive power and flexibility of deep learning.

LIDWT aggregates the outputs  of all encoder streams ($\{H^{(\omega)}\}_{\omega=1}^{\Omega+1}$) by
\begin{equation}
\begin{aligned}
&b^{(\omega-1)}_i=D^{(\omega)}_{L}(k)a^{(\omega)}_k+D^{(\omega)}_{H}(k)b^{(\omega)}_k,\quad i=2k;\\
&b^{(\omega-1)}_i=A^{(\omega)}_{L}(k)a^{(\omega)}_k+A^{(\omega)}_{H}(k)b^{(\omega)}_k,\quad i=2k+1;\\
&b^{(\Omega)}_k=H^{(\Omega+1)}_{k,:},\quad a^{(\omega)}_k=H^{(\omega)}_{k,:},\quad\omega=1,...,\Omega.
\end{aligned}
\end{equation}
$\{D_L,D_H,A_L,A_H\}$ are  the learnable parameters that serve as reconstruction coefficients in IDWT.

\subsection{Loss function and Train Strategy}

We use Mean Absolute Error (MAE) as the loss function,
\begin{equation}
L_{pre}=\frac{1}{T}\sum_{i=1}^T|\hat{X}_{t+i}-X_{t+i}|,
\end{equation}
where $X$ is the ground truth and $\hat{X}$ is prediction result.

Adaptive moment estimation is used as the basic training algorithm.
Furthermore, in order to alleviate performance degradation caused by discrepancy between the input distributions of training and testing, we adopt scheduled sampling method \cite{li2017dcrnn},
and to reduce the time and memory costs of GCRN, we use curriculum learning strategy \cite{li2023dynamic}.

\begin{table*}[tb]
\centering
\caption{Comparison on METR-LA dataset }\label{ex1}
\begin{tabular}{*{10}{c}}
  \toprule
  \multirow{2}*{Model} & \multicolumn{3}{c}{Horizon 3} & \multicolumn{3}{c}{Horizon 6} & \multicolumn{3}{c}{Horizon 12}\\
  \cmidrule(lr){2-4}\cmidrule(lr){5-7}\cmidrule(lr){8-10}
  & MAE & RMSE & MAPE & MAE & RMSE & MAPE & MAE & RMSE & MAPE \\
  \midrule
  DCRNN & 2.77 & 5.38 & 7.30\% & 3.15 & 6.45 & 8.80\% & 3.60 & 7.59 & 10.50\% \\
  \midrule
  STGCN & 2.88 & 5.74 & 7.62\% & 3.47 & 7.24 & 9.57\% & 4.59 & 9.40 & 12.70\% \\
  \midrule
  Graph WaveNet & 2.69 & 5.15 & 6.90\% & 3.07 & 6.22 & 8.37\% & 3.53 & 7.37 & 10.01\% \\
  \midrule
  AGCRN & 2.87 & 5.58 & 7.70\% & 3.23 & 6.58 & 9.00\% & 3.62 & 7.51 & 10.38\% \\
  \midrule
  MTGNN & 2.69 & 5.18 & 6.86\% & 3.05 & 6.17 & 8.19\% & 3.49 & 7.23 & 9.87\% \\
  \midrule
  WavGCRN & \textbf{2.67} & \textbf{5.13} & \textbf{6.76\%} & \textbf{3.05} & \underline{6.22} & \textbf{8.15\%} & \underline{3.50} & 7.38 & \textbf{9.81\%} \\
  \bottomrule
\end{tabular}
\end{table*}

\begin{table*}[htb]
\centering
\caption{Comaprison on PEMS-BAY dataset}\label{ex2}
\begin{tabular}{*{10}{c}}
  \toprule
  \multirow{2}*{Model} & \multicolumn{3}{c}{Horizon 3} & \multicolumn{3}{c}{Horizon 6} & \multicolumn{3}{c}{Horizon 12}\\
  \cmidrule(lr){2-4}\cmidrule(lr){5-7}\cmidrule(lr){8-10}
  & MAE & RMSE & MAPE & MAE & RMSE & MAPE & MAE & RMSE & MAPE \\
  \midrule
  DCRNN & 1.38 & 2.95 & 2.90\% & 1.74 & 3.97 & 3.90\% & 2.07 & 4.74 & 4.90\% \\
  \midrule
  STGCN & 1.36 & 2.96 & 2.90\% & 1.81 & 4.27 & 4.17\% & 2.49 & 5.69 & 5.79\% \\
  \midrule
  Graph WaveNet & 1.30 & 2.74 & 2.73\% & 1.63 & 3.70 & 3.67\% & 1.95 & 4.52 & 4.63\% \\
  \midrule
  AGCRN & 1.37 & 2.87 & 2.94\% & 1.69 & 3.85 & 3.87\% & 1.96 & 4.54 & 4.64\% \\
  \midrule
  MTGNN & 1.32 & 2.79 & 2.77\% & 1.65 & 3.74 & 3.69\% & 1.94 & 4.49 & 4.53\% \\
  \midrule
  WavGCRN & \textbf{1.30} & \textbf{2.73} & \textbf{2.73\%} & \underline{1.64} & 3.75 & \underline{3.68\%} & \textbf{1.94} & 4.58 & 4.62\% \\
  \bottomrule
\end{tabular}
\end{table*}

\subsection{Graph Learning}

In most of GCN based traffic metrics forecasting methods, either the graph is directly derived from the reachable shortest distances between sensor locations in road network or learned from traffic data during training procedure. In this paper we propose a new graph learning method that combines these two kinds of methods together, which simultaneously utilize both physics-informed and data-informed information.

\begin{equation}
 A = \gamma A_{data} + (1-\gamma) A_{road}.
\end{equation}
$A_{data}$ is the  learned graph and $A_{dist}$ is prior graph directly derived from road network. $\gamma$ considers the balance between two graphs.
The iterative updating scheme is used to balance the current learned graph and the last learned graph, enabling the graph learning smooth.

Most of graph learning methods employed in previous work like \cite{wu2020connecting} and \cite{li2023dynamic} are based on the similarity between node features.
However, node feature similarity-based graph cannot explicitly represent causal relationship and manifold structure.
As a consequence, we turn to optimization-based graph learning algorithm,  named NOTEARS in \cite{zheng2018dags}.

NOTEARS estimates the structure of a directed acyclic graph (DAG) $A$ by
\begin{equation}
\begin{aligned}
&\mbox{minimize }\quad\text{Score}(A;X) = \frac{1}{2B}\|X-XA\|_F^2+\lambda\|A\|_1, \\
&\mbox{subject to }\quad \text{tr}(e^{A\otimes A}) - N = 0.
\end{aligned}
\end{equation}
 $\text{Score}$ function evaluates whether the graph structure effectively reflects the node causal relationships and the constraint holds iff there is no cycle in the graph $A$. We use L-BFGS \cite{byrd1995limited} algorithm to solve this optimization problem.

Since the spatial correlations in different frequency-domains are also different, e.g., the graph underlying low-frequency signal component might be quite sparser than the graph of high-frequency component.
The different graphs are learned separately for each input stream.

\section{Experiment}
\label{sec:ex}

Two popular real-world datasets are used in our experiments.
\textbf{\bf}{METR-LA} was gathered from a network of $207$ loop detectors strategically positioned throughout the Los Angeles highway system, the time spans from 03/01/2012 to 06/30/2012  \cite{jagadish2014big}.
\textbf{\bf}{PEMS-BAY} was acquired from $325$ sensors spanning the extensive Bay Area landscape, and covers a period from 01/01/2017 to 05/31/2017   \cite{li2017dcrnn}. Both of them provide average traffic flow speeds in 5 minutes in miles per hour (mph).  All experiments are carried out using a system based on the NVIDIA DGX A100 platform. Our codebase can be found at \url{https://github.com/qqian99/WavGCRN}. 


WavGCRN is compared with several state-of-the-art traffic prediction approaches, including DCRNN \cite{li2017dcrnn}, STGCN \cite{yu2017spatio}, Graph WaveNet \cite{wu2019graph}, AGCRN \cite{bai2020adaptive}, and MTGNN \cite{wu2020connecting}.
Three popular criterions Mean Absolute Error (MAE), Root Mean Squared Error (RMSE), and Mean Absolute Percentage Error (MAPE) are used for evaluating forecasting performance.
The experiments are set  by 15 minutes (horizon 3), 30 minutes (horizon 6), and 1 hour (horizon 12)
ahead forecasting based on 1-hour historical input data from the META-LA and the PEMS-BAY datasets.

Results are outlined in Tables \ref{ex1} and \ref{ex2} respectively. Apart from WavGCRN, the results of all other methods are directly copied from their papers and \cite{li2023dynamic}. The \textbf{best} and \underline{second best} results are highlighted. It can be seen that WavGCRN exhibits superior performance as a whole compared with other models. Especially, our approach demonstrates a prominent advantage in short-range horizon prediction. Notice that traffic condition in METR-LA is more complex than in PEMS-BAY, as Los Angeles is renowned for its intricate traffic dynamics. Thus the forecasting on METR-LA is also more difficult than on PEMS-BAY, and the superiority of our method on METR-LA is more significant. 

\section{Conclusion}

In this paper, a novel MSA-based and DL-based hybrid method WavGCRN, is proposed for traffic forecasting, which is a DWT-inspired encoder-decoder network architecture with GCRN and graph learning. Compared with the current DL-based traffic forecasting methods, our model is able to better explore the multiscale spatiotemporal relationship buried among the traffic metrics data and has more concrete physical interpretability. The proposed method can be used for other application scenarios in which time series signals are observed on associated sensors, such as Internet of Things and social networks.

\section*{Acknowledgments}
This research was supported by the National Science Foundation through the MSGI Internship Program and the U.S. Department of Energy, Office of Science, Advanced Scientific Computing Research, through the SciDAC-RAPIDS2 institute under Contract DE-AC02-06CH11357. Additionally, it utilized resources from the Argonne Leadership Computing Facility under contract DE-AC02-06CH11357.


\bibliographystyle{IEEEbib}
\bibliography{refs}

\end{document}